\documentclass{article}
\usepackage[accepted]{eiml_icml2026}

\usepackage{amsmath,amssymb,amsthm,mathtools}
\usepackage{bm}
\usepackage{booktabs}
\usepackage{multirow}
\usepackage{enumitem}
\usepackage{graphicx}
\usepackage{microtype}
\usepackage{hyperref}

\hypersetup{colorlinks=true,
linkcolor=blue!70!black,citecolor=blue!50!black,urlcolor=blue!60!black}

\newtheorem{proposition}{Proposition}
\theoremstyle{remark}
\newtheorem{remark}{Remark}

\newcommand{\Dtrain}{\mathcal{D}_{\mathrm{tr}}}
\newcommand{\Dcal}{\mathcal{D}_{\mathrm{cal}}}
\newcommand{\R}{\mathbb{R}}
\newcommand{\E}{\mathbb{E}}

\newcommand{\X}{\mathcal{X}}
\newcommand{\indic}{\mathbb{I}}
\newcommand{\hth}{\widehat{\theta}}
\newcommand{\hmu}{\widehat{\mu}}
\newcommand{\hsi}{\widehat{\sigma}}
\newcommand{\hq}{\widehat{q}}

\icmltitlerunning{Conformal Bayes under Label Shift}

\begin{document}

\twocolumn[
\icmltitle{Conformal Bayes under Label Shift:\\
  Post-Hoc Calibration vs.\ In-Training Adaptation}

\begin{icmlauthorlist}
  \icmlauthor{Seungjin Choi}{croid}
\end{icmlauthorlist}
\icmlaffiliation{croid}{CROID Research and aSSIST University, Seoul, Korea}
\icmlcorrespondingauthor{Seungjin Choi}{seungjin.choi.mlg@gmail.com}
\icmlkeywords{conformal prediction, Bayesian inference, label shift,
  posterior tilting, predictive tilting, epistemic uncertainty}
\vskip 0.3in]

\printAffiliationsAndNotice{}

\begin{abstract}
Conformal Bayes combines Bayesian posterior predictives with conformal calibration 
to produce prediction sets that are both statistically valid and geometrically efficient.
We study conformal Bayes under label shift from a unified perspective, identifying two complementary 
approaches that restore nominal target-domain coverage through
importance-weighted conformal calibration but operate through independent mechanisms.
\emph{Post-hoc calibration} tilts the posterior predictive toward the target domain and corrects the 
conformal threshold via an importance-weighted quantile, leaving the parameter posterior unchanged.
\emph{In-training adaptation} tilts the parameter posterior itself to the target domain, 
producing a corrected predictive whose highest predictive density region serves as the
highest predictive density (HPD)-based prediction set under the fitted target predictive; efficiency is model-dependent
and does not imply finite-sample conditional optimality.
Two controlled experiments isolate the regime-dependence of each strategy:
in the low-dimensional, well-estimated regime Strategy~A produces the narrowest valid intervals,
while in the high-dimensional, underdetermined regime Strategy~B achieves up to $43\%$
width reduction at unchanged coverage, under the stated source-sampling and label-shift assumptions.
\end{abstract}

\section{Introduction}
 
Conformal prediction \cite{VovkV2005book,PapadopoulosH2002ecml,ShaferG2008jmlr,BarberRF2023aos,AngelopoulosAN2023ftml} 
is a model-agnostic framework
that constructs prediction sets with finite-sample coverage guarantees from exchangeable data, 
free of any distributional assumption.
\emph{Conformal Bayes} \cite{WassermanL2011ss,FongE2021neurips} instantiates this framework 
by using the negative log-predictive density as a Bayesian nonconformity score,
encoding both aleatoric and \emph{epistemic} uncertainty.
Under \emph{label shift} \cite{SaerensM2002neco,LiptonZ2018icml,AzizzadenesheliK2019iclr,AlexandariAM2020icml},
the target label marginal and the source label marginal are not equal,
i.e., $p_t(y)\neq p_s(y)$ while $p_t(x|y)=p_s(x|y)$.
In such a case, calibration and test data are no longer exchangeable under the same measure, and
prediction sets constructed from source data fail to cover target test points at the nominal level
\cite{PodkopaevA2021uai,LeeHS2025reliableML}.
This arises naturally in lead-optimization, where training data are
enriched for high-activity compounds but predictions are required over the full chemical library.
A separate difficulty appears in high-dimensional, data-limited regimes, where large posterior
uncertainty can make the geometry of a tilted predictive score unstable.

In this paper we consider two strategies derived from the posterior predictive.
The geometry of the resulting prediction set directly reflects the model's 
\emph{epistemic uncertainty} about the parameters $\theta$: inputs whose feature 
representations lie far from the training distribution receive wider prediction sets, 
automatically reflecting genuine uncertainty rather than a fixed inflation.
Under label shift, this uncertainty may be miscalibrated for the target domain, 
motivating the two strategies we study: (1) post-hoc calibration; (2) in-training adaptation.

\emph{Post-hoc calibration} tilts the posterior predictive toward the target domain and corrects the 
conformal threshold via an importance-weighted quantile, leaving the parameter posterior unchanged.
\emph{In-training adaptation} tilts the parameter posterior itself to the target domain, 
producing a corrected predictive whose highest predictive density region serves as the
HPD-based prediction set under the fitted target predictive; efficiency is model-dependent
and does not imply finite-sample conditional optimality.
The two strategies differ in \emph{where} the label shift correction is applied: 
at the predictive level (post-hoc) or at the parameter level (in-training), 
and this distinction has concrete consequences for interval geometry
and computational requirements.
We study these two approaches analytically in the case of Bayesian linear regression (Gaussian)
and demonstrate their behaviors via two controlled experiments.

\paragraph{Contributions:}
(1) A unified conformal-Bayes view of label shift separating weighted calibration for coverage from predictive/posterior tilting for score geometry, with a closed-form tilted posterior for Gaussian heads;
(2) A head-to-head comparison identifying when each strategy is preferable;
(3) Synthetic experiments in two controlled regimes showing when predictive tilting is preferable and when posterior tilting gives narrower intervals at the same nominal coverage.

\section{Background}\label{sec:background}

\subsection{Model and Label Shift}

Let $\Dtrain=\{(x_i,y_i)\}_{i=1}^n$ with $(x_i,y_i)\overset{\text{i.i.d.}}{\sim}p_s$.
We use a frozen backbone $\phi:\X\to\R^d$ and Gaussian linear head:
\begin{equation}
y\mid x,\theta\;\sim\;\mathcal{N}(\theta^\top\phi(x),\,\sigma^2),
\qquad
\theta\;\sim\;\mathcal{N}(0,\,\tau^2 I_d).
\label{eq:model}
\end{equation}
The model is discriminative: $\theta$ parameterizes $p(y|x,\theta)$
only, and the covariate marginal $p(x)$ does not depend on $\theta$
since the backbone $\phi$ is frozen.
The source posterior is $\pi_s(\theta|\Dtrain)=\mathcal{N}(\hth_s,\Sigma_\theta)$
by conjugacy, with
$\Sigma_\theta=(\tau^{-2}I_d+\sigma^{-2}\Phi^\top\!\Phi)^{-1}$
and $\hth_s=\Sigma_\theta\Phi^\top y/\sigma^2$.

Label shift is modeled by exponential tilting:
\begin{equation}
w(y)= \frac{p_t(y)}{p_s(y)}=\frac{\exp(\beta y)}{Z_w},
\quad \beta\in\R,
\label{eq:tilt}
\end{equation}
where $\quad Z_w=\int\exp(\beta y)\,p_s(y)\,dy$.
Under~\eqref{eq:tilt} the target marginal is $p_t(y)\propto p_s(y)\exp(\beta y)$,
and the target conditional is
$p_t(y|x,\theta)=\mathcal{N}(\theta^\top\phi(x)+\beta\sigma^2,\sigma^2)$
(see Appendix~\ref{app:stratB}).
For $\beta>0$ the target label distribution is shifted toward higher
values; for $\beta<0$ toward lower values.

\subsection{Conformal Bayes}

Standard split conformal prediction \cite{PapadopoulosH2002ecml,AngelopoulosAN2023ftml} builds
a prediction set from a nonconformity score $s(x,y)$ and a held-out
calibration set $\Dcal=\{(x_i,y_i)\}_{i=1}^m\sim p_s$.
The $(1-\alpha)$-quantile of the calibration scores,
$\hq = \mathrm{Quantile}_{1-\alpha}\{s(x_i,y_i)\}_{i=1}^m$,
defines $C(x)=\{y:s(x,y)\leq\hq\}$, which satisfies
$\Pr(y_{\mathrm{test}}\in C(x_{\mathrm{test}}))\geq 1-\alpha$
whenever calibration and test are exchangeable.

\emph{Conformal Bayes} \cite{MelluishT2001ecml,WassermanL2011ss,FongE2021neurips} uses the Bayesian posterior
predictive $p(y|x,\Dtrain)=\int p(y|x,\theta)\,\pi_s(\theta|\Dtrain)\,d\theta$
as the basis for the nonconformity score.
The natural choice is the negative log-predictive density (NLPD):
\begin{equation}
s(x,y) = -\log p(y | x,\Dtrain).
\label{eq:nll-score}
\end{equation}
The resulting prediction set $C(x)=\{y:p(y|x,\Dtrain)\geq e^{-\hq}\}$
is the HPD region of $p(\cdot|x,\Dtrain)$: the smallest set
concentrating the most predictive probability mass.
By the Neyman--Pearson lemma, HPD regions are minimum-volume sets
at a given predictive probability level under the fitted model density;
this is a model-dependent efficiency property and does not imply
finite-sample conditional optimality of the conformal set.
Alternative nonconformity scores such as conformalized quantile
regression \cite{RomanoY2019neurips} and distribution-free predictive
inference \cite{LeiJ2018jasa} achieve local adaptivity through different
mechanisms; Conformal Bayes achieves adaptivity through the posterior
predictive variance $\hsi^2(x)$.

\subsection{Assumptions}

Both strategies operate under assumptions (A1) and (A2).
Strategy~A additionally requires (A3).

\textbf{(A1) Label shift (conditional invariance).}
The conditional distribution of features given the label is invariant
across domains, $p_t(x|y)=p_s(x|y)$, while the marginal label
distribution shifts, $p_t(y)\neq p_s(y)$.
Because the backbone $\phi$ is frozen and $\Dtrain$ is drawn from
$p_s$ and fixed, this conditional invariance extends to the learned
predictive mechanism: $p_t(x|y,\Dtrain)=p_s(x|y,\Dtrain)$.

\textbf{(A2) Absolute continuity and density ratio.}
The source marginal has support wherever the target marginal does:
$p_s(y)>0$ for all $y$ where $p_t(y)>0$, ensuring $w(y)=p_t(y)/p_s(y)$
is well-defined.
We model $w(y)$ by exponential tilting~\eqref{eq:tilt}, the
maximum-entropy family subject to a mean constraint, with $\beta$ treated as a known parameter.

\textbf{(A3) Source representativeness (Strategy~A only).}
The training data are representative of the source population:
$p_s(y|\Dtrain)\approx p_s(y)$.
This holds in the limit $n\to\infty$ by posterior concentration
and is mild for reasonably large $n$.
Strategy~B does not require (A3): its corrected posterior is derived
entirely from the per-observation likelihood ratio $p_t(y|x,\theta)/p_s(y|x,\theta)$,
which is a $\theta$-level calculation that never invokes $p_s(y|\Dtrain)$.
Both experiments use unbiased training data, so (A3) holds in both cases.

Under (A1), calibration data $(x_i,y_i)\overset{\text{i.i.d.}}{\sim}p_s$
are exchangeable with the test point under the importance-weighted
measure $w(y)\cdot p_s$, which is the condition for the
importance-weighted conformal guarantee \citep{TibshiraniR2019neurips,BarberRF2023aos}.

\section{Conformal Bayes under Label Shift}\label{sec:methods}

\subsection{Strategy A: Post-Hoc Calibration via Predictive Tilting}

Strategy~A leaves $\pi_s(\theta|\Dtrain)$ unchanged.
Under assumptions (A1)--(A3), the target posterior predictive
equals the tilted source predictive (proof in
Appendix~\ref{app:stratA}):
\begin{eqnarray}
p^{\mathrm{A}}(y|x,\Dtrain)
& = & p_t(y|x,\Dtrain) \nonumber \\
& = & \frac{p_s(y|x,\Dtrain)\,w(y)}{Z(x)},
\label{eq:pred-tilt}
\end{eqnarray}
where $Z(x)=\int p_s(y'|x,\Dtrain)\,w(y')\,dy'$.
The proof works entirely at the predictive level via Bayes' theorem
and the label shift assumption, without any marginalization over
$\theta$.
Strategy~A uses the negative log-density of $p^{\mathrm{A}}$ as the
nonconformity score:
\begin{eqnarray}
s^{\mathrm{A}}(x,y) & = & -\log p^{\mathrm{A}}(y | x,\Dtrain) \nonumber \\
& = & -\log \frac{p_s(y\mid x,\Dtrain)\,w(y)}{Z(x)}.
\label{eq:sA}
\end{eqnarray}
For the Gaussian head, $s^{\mathrm{A}}$ is the NLPD of
$\mathcal{N}(\hmu_s(x)+\beta\hsi^2(x),\,\hsi^2(x))$.
The conformal quantile is importance-weighted:
\begin{equation}
\hq^{\mathrm{A}}=\inf\!\Bigl\{q:
\tfrac{\sum_i w(y_i)\indic\{s_i^{\mathrm{A}}\leq q\}}
      {\sum_j w(y_j)}\geq 1-\alpha\Bigr\},
\label{eq:qA}
\end{equation}
giving $C^{\mathrm{A}}(x)=\{y:s^{\mathrm{A}}(x,y)\leq\hq^{\mathrm{A}}\}$.
The exact candidate-weighted version satisfies the finite-sample weighted conformal guarantee
\cite{TibshiraniR2019neurips}.
The practical quantile~\eqref{eq:qA}, which omits the candidate test weight, is the
large-calibration approximation used in our experiments and is asymptotically valid
as $n_{\mathrm{cal}}\to\infty$ (see Remark~\ref{rem:testweight}).
In practice, the normalizing constant $Z_w$ in $w(y)=\exp(\beta y)/Z_w$
cancels in the ratio $\sum_i w(y_i)\indic\{\cdot\}/\sum_j w(y_j)$,
so Algorithm~\ref{alg:main} computes $w_i\leftarrow\exp(\widehat\beta\,y_i)$ without $Z_w$.
In the theory and experiments below we set $\widehat\beta=\beta$ (oracle);
the notation $\widehat\beta$ in Algorithm~\ref{alg:main} highlights the plug-in form
used when $\beta$ is estimated in practice.

\begin{remark}[Test-point weight and finite-sample validity]\label{rem:testweight}
The exact finite-sample guarantee \cite{TibshiraniR2019neurips} requires including
$w(y_{\mathrm{test}})$ in the weighted CDF, making the prediction set implicitly defined as
$\{y : \sum_i w(y_i)\indic\{s_i\geq s(x_{\mathrm{test}},y)\}+w(y) > \alpha(\sum_i w(y_i)+w(y))\}$.
Equation~\eqref{eq:qA} omits this test-point weight — the standard large-$n_{\mathrm{cal}}$
approximation, asymptotically valid as $n_{\mathrm{cal}}\to\infty$.
\end{remark}

\subsection{Strategy B: In-Training Adaptation via Posterior Tilting}

Strategy~B corrects the parameter posterior before calibration.
A naive reweighting $\pi_s(\theta|\Dtrain)\cdot\prod_i w(y_i)$ has no
effect, since $w(y_i)$ is $\theta$-free and cancels in normalization.
The correct update uses the per-observation likelihood ratio
$p_t(y_i|x_i,\theta)/p_s(y_i|x_i,\theta)=w(y_i)/Z_\theta(x_i)$
(derived in Appendix~\ref{app:stratB}), giving:
\begin{equation}
\pi_t(\theta|\Dtrain)
\;\propto\;
\pi_s(\theta|\Dtrain)\cdot\prod_{i=1}^n Z_\theta(x_i)^{-1}.
\label{eq:post-t}
\end{equation}
This is a likelihood-ratio-adjusted pseudo-posterior, not the posterior
one would obtain from genuine target training data: $\Dtrain$ is drawn
from $p_s$, so $\pi_t(\theta|\Dtrain)$ treats the source observations
as if they were informative under the target conditional likelihood.
We use the term \emph{target-tilted posterior} to emphasize that this is
a likelihood-ratio-corrected posterior induced by the assumed tilting model,
rather than a posterior obtained from labeled target training data.
Under label shift this is a principled approximation — the conditional
$p(x|y)$ is invariant (Assumption~A1), so the source inputs $x_i$
remain valid features — but the result should be interpreted as a
corrected pseudo-posterior rather than a fully Bayesian target posterior.

\begin{proposition}[Closed-Form Target-Tilted Posterior]\label{prop:map}
Under~\eqref{eq:model} and exponential tilting~\eqref{eq:tilt},
the target-tilted posterior is Gaussian:
$\pi_t(\theta|\Dtrain)=\mathcal{N}(\hth_t,\Sigma_\theta)$ with
\begin{equation}
\hth_t = \hth_s - \beta n\Sigma_\theta\bar\phi,
\qquad
\bar\phi = n^{-1}\textstyle\sum_{i=1}^n\phi(x_i),
\label{eq:map}
\end{equation}
and unchanged covariance $\widehat\Sigma_t=\Sigma_\theta$.
\end{proposition}

\noindent(Proof in Appendix~\ref{app:stratB}.)

The corrected predictive is
$p^{\mathrm{B}}(y|x)=\mathcal{N}(\hmu_t(x),\hsi^2(x))$
with $\hmu_t(x)=\hth_t^\top\phi(x)+\beta\sigma^2$ and
$\hsi^2(x)=\sigma^2+\phi(x)^\top\Sigma_\theta\phi(x)$ (unchanged).
Strategy~B uses $s^{\mathrm{B}}(x,y)=-\log p^{\mathrm{B}}(y|x)$ with the same
weighted quantile~\eqref{eq:qA}, giving
$C^{\mathrm{B}}(x)=\{y:p^{\mathrm{B}}(y|x)\geq e^{-\hq^{\mathrm{B}}}\}$.

\if(0)
{\bf Structural comparison of the two interval centers:}
\begin{itemize}
\item
\text{Predictive tilting}
\begin{equation}
  \hmu_s(x)+\beta\sigma^2
  +\underbrace{\beta\phi(x)^\top\Sigma_\theta\phi(x)}_{\text{epistemic uncertainty shift}}.
  \label{eq:cA}
\end{equation}
\item
\text{Posterior tilting}
\begin{equation}
  \hmu_s(x)+\beta\sigma^2
  -\underbrace{\beta n\phi(x)^\top\Sigma_\theta\bar\phi}_{\text{training-similarity shift}}.
  \label{eq:cB}
\end{equation}
\end{itemize}
\fi

\paragraph{Structural comparison of the two interval centers.}
\begin{align}
\text{Predictive tilting:}&\quad
  \hmu_s(x)+\beta\sigma^2
  +\underbrace{\beta\phi(x)^\top\Sigma_\theta\phi(x)}_{\text{epistemic uncertainty shift}},
  \label{eq:cA}\\
\text{Posterior tilting:}&\quad
  \hmu_s(x)+\beta\sigma^2
  -\underbrace{\beta n\phi(x)^\top\Sigma_\theta\bar\phi}_{\text{training-similarity shift}}.
  \label{eq:cB}
\end{align}
Both share the $+\beta\sigma^2$ offset from the label-shift-induced
change in the conditional mean.
Predictive tilting adds a further shift proportional to the
epistemic uncertainty $\phi(x)^\top\Sigma_\theta\phi(x)$, which is
larger for out-of-distribution inputs where the posterior over $\theta$
is diffuse.
This is a direct consequence of using the full posterior predictive
as the score: the model's uncertainty about the parameters is
encoded in the prediction set geometry, not discarded as in a
residual-based score.
Posterior tilting shifts by the similarity of the test point to
the training distribution, applying zero additional correction to
out-of-distribution inputs.

Both strategies use the full predictive variance $\hsi^2(x)$ in the NLPD score,
so epistemic uncertainty governs interval width in both cases.

The two strategies address the same target predictive distribution through
different correction mechanisms: the importance-weighted (IW) quantile~\eqref{eq:qA}
is the principled source of validity for both, while the choice of score function
(predictive tilting vs.\ posterior tilting) determines geometric efficiency.
Neither mechanism subsumes the other.

\if(0)
\begin{remark}[Epistemic uncertainty and prediction set geometry]\label{rem:epistemic}
Standard split conformal prediction uses the absolute residual
$|y-\hat f(x)|$ as the nonconformity score.
This score is invariant to input-specific predictive uncertainty:
an in-distribution input and an out-of-distribution input with the same
residual receive identical scores, and hence equally wide prediction sets.
The NLPD score $s(x,y)=-\log p(y|x,\Dtrain)$ incorporates the
predictive variance $\hsi^2(x)=\sigma^2+\phi(x)^\top\Sigma_\theta\phi(x)$,
so inputs for which the parameter posterior is diffuse --- inputs
whose feature representations lie far from the training set ---
naturally attract wider prediction sets at the same coverage level.
Under label shift, this epistemic sensitivity is further amplified:
the epistemic term $\beta\hsi^2(x)$ shifts the interval center
for Strategy~A, applying a larger correction precisely where the
model is least certain about the target label.
\end{remark}
\fi

\subsection{Algorithms and Comparison}\label{sec:compare}

Both strategies share the same pipeline and differ only in the score
function (Step~2 in Algorithm \ref{alg:main}).

\begin{algorithm}[ht!]
\caption{Conformal Bayes under Label Shift}
\label{alg:main}
\begin{algorithmic}[1]
\REQUIRE Training set $\Dtrain$, calibration set $\Dcal$,
  label shift parameter $\beta$,
  backbone $\phi(\cdot)$, coverage level $1-\alpha$,
  strategy choice (\texttt{A} or \texttt{B})
\ENSURE Prediction set $C(x_{\mathrm{test}})$

\STATE \textbf{// Step 1: Source Bayesian linear regression}
\STATE Compute $\Sigma_\theta\!\leftarrow\!
  (\tau^{-2}I_d+\sigma^{-2}\Phi^\top\!\Phi)^{-1}$,\;
  $\hth_s\!\leftarrow\!\Sigma_\theta\Phi^\top y/\sigma^2$

\STATE \textbf{// Step 2: Compute score function}
\IF{\texttt{Strategy A}}
  \STATE $s^{\mathrm{A}}(x,y)\leftarrow
    \text{NLPD}\bigl(y;\;\hmu_s(x)+\widehat\beta\hsi^2(x),\;\hsi^2(x)\bigr)$
\ELSE[\texttt{Strategy B}]
  \STATE $\hth_t\leftarrow\hth_s
    -\widehat\beta\,\Sigma_\theta\!\sum_{i=1}^n\phi(x_i)$
    \hfill\COMMENT{$O(nd)$, Eq.~\eqref{eq:map}}
  \STATE $s^{\mathrm{B}}(x,y)\leftarrow
    \text{NLPD}\bigl(y;\;\hmu_t(x),\;\hsi^2(x)\bigr)$
    \quad where $\hmu_t(x)=\hth_t^\top\phi(x)+\widehat\beta\sigma^2$
\ENDIF

\STATE \textbf{// Step 3: Score calibration set}
\FOR{each $(x_i,y_i)\in\Dcal$}
  \STATE Compute $s_i$ using the chosen score function
\ENDFOR

\STATE \textbf{// Step 4: Importance-weighted quantile}
\STATE Compute $w_i\leftarrow\exp(\widehat\beta\,y_i)$ for each
  $(x_i,y_i)\in\Dcal$
\STATE $\hq\leftarrow\inf\!\bigl\{q:
  {\textstyle\sum_i} w_i\indic\{s_i\leq q\}/
  {\textstyle\sum_j} w_j\geq 1-\alpha\bigr\}$
  \hfill\COMMENT{with $+\infty$ sentinel \cite{TibshiraniR2019neurips}}

\STATE \textbf{// Step 5: Prediction set}
\STATE $C(x_{\mathrm{test}})\leftarrow\{y : s(x_{\mathrm{test}},y)\leq\hq\}$
\RETURN $C(x_{\mathrm{test}})$
\end{algorithmic}
\end{algorithm}

\if(0)
\paragraph{Complexity.}
Step~1: $O(nd^2+d^3)$ (posterior computation, once).
Step~3B: $O(nd)$ (one matrix-vector multiply beyond Strategy~A).
Steps~4--5: $O(md)$ where $m=|\Dcal|$.
Total extra cost of Strategy~B over Strategy~A: one $O(nd)$ multiply.
\fi


{\bf When predictive tilting is preferable:}
When the model is low-dimensional and well-estimated ($n_{\mathrm{tr}}\gg d$),
the epistemic uncertainty $\hsi^2(x)$ is small and stable.
Strategy~A's tilted score is well-aligned with the true target predictive
and produces the narrowest valid intervals (Experiment~1, Section~\ref{sec:exp1}).
It also requires no access to the training pipeline and applies to any pre-trained model.

{\bf When posterior tilting is preferable:}
When the model is high-dimensional and underdetermined ($n_{\mathrm{tr}}\lesssim d$),
the large epistemic uncertainty $\hsi^2(x)$ makes Strategy~A's
tilted score noisy and the IW quantile inflated.
Strategy~B's MAP correction shifts the predictive mean by a stable vector
independent of $x$'s epistemic uncertainty, keeping calibration scores compact
and producing substantially narrower intervals — up to $43\%$ reduction at
$\beta=0.6$ (Experiment~2, Section~\ref{sec:exp2}) —
with all assumptions (A1)--(A3) satisfied.
The trade-off is sensitivity to $\widehat\beta$ and the requirement for training pipeline access.


\section{Experiments}\label{sec:exp}

\subsection{Synthetic Data Generation}\label{sec:dgp}

Label shift is induced by exponential tilting~\eqref{eq:tilt} with
$\beta\in\{0,0.1,\ldots,0.6\}$.
The target joint is $p_t(x,y)\propto p_s(x,y)\exp(\beta y)$;
test points are drawn by resampling pairs $(x,y)$ from a large source
pool with probabilities proportional to $\exp(\beta y)$.
Throughout, $\beta$ is treated as a known parameter.

\paragraph{Experiment~1 (low-dimensional, unbiased training).}
We use $d=5$, features $\phi(x)=x+\mathbf{1}_d$,
$x\sim\mathcal{N}(0,\Lambda)$,
$\Lambda=\mathrm{diag}(1,2,3,4,5)$,
$\sigma^2=1$, $\tau^2=2$,
$\theta_{\mathrm{true}}=(0.2,-0.1,0.3,-0.2,0.1)^\top$.
Training and calibration are both uniform draws from $p_s$
($n_{\mathrm{tr}}=2000$, $n_{\mathrm{cal}}=2000$).
This is the \emph{low-dimensional, well-estimated} regime:
$\hth_s\approx\theta_{\mathrm{true}}$ and $\hsi^2(x)$ is moderate.

\paragraph{Experiment~2 (high-dimensional, underdetermined).}
We use $d=50$, features $\phi(x)=x+\mathbf{1}_d$,
$x\sim\mathcal{N}(0,\Lambda_d)$,
$\Lambda_d=\mathrm{diag}(1/d,\ldots,d/d)$,
$\sigma^2=1$, $\tau^2=2$,
$\theta_{\mathrm{true}}\sim\mathcal{N}(0,d^{-1}I_d)$ (fixed across seeds).
Training and calibration are uniform draws from $p_s$
($n_{\mathrm{tr}}=50$, $n_{\mathrm{cal}}=2000$).
All assumptions (A1)--(A3) are satisfied: training is unbiased,
the model is correctly specified.
This is the \emph{underdetermined} regime: $n_{\mathrm{tr}}\ll d$,
so the posterior covariance $\Sigma_\theta$ is large and
$\hsi^2(x)\approx 6.4$ averaged over test points.

Table~\ref{tab:expdesign} summarizes both experiments.
All results are averaged over $n_{\mathrm{seeds}}=300$ independent random seeds;
$n_{\mathrm{te}}=2000$ throughout.

\begin{table}[ht]
\centering
\caption{Experimental design. Both experiments use unbiased training
and calibration from $p_s$; they differ in dimension and training-set size.}
\label{tab:expdesign}
\smallskip\small
\setlength{\tabcolsep}{4pt}
\begin{tabular}{lcccc}
\toprule
 & \textbf{Train} & \textbf{Cal.} & \textbf{Test/Target} & \textbf{$d$, $n_{\mathrm{tr}}$} \\
\midrule
Exp.~1 & $p_s$ & $p_s$ & $p_t\propto p_s e^{\beta y}$ & $5$,\ $2000$ \\
Exp.~2 & $p_s$ & $p_s$ & $p_t\propto p_s e^{\beta y}$ & $50$,\ $50$ \\
\bottomrule
\end{tabular}
\end{table}

\subsection{Methods in Comparison}

Four methods are evaluated:
\begin{enumerate}
\item \textbf{CB (unweighted $\widehat{q}$)}: source NLPD score,
unweighted quantile — the naive baseline ignoring label shift.
\item \textbf{CB (IW $\widehat{q}$)}: source NLPD score, IW quantile —
equivalent to a non-Bayesian IW conformal predictor \cite{TibshiraniR2019neurips}
using the Bayesian NLPD score; serves as the direct non-Bayesian baseline.
\item \textbf{Strategy~A} (predictive tilting): tilted NLPD score, IW quantile.
\item \textbf{Strategy~B} (posterior tilting): corrected NLPD score, IW quantile.
\end{enumerate}
Methods 2--4 all use Algorithm~\ref{alg:main}; they differ only in the
score function at Step~2.
Experiment~2 does not model lead-optimization enrichment; instead, it
isolates a separate regime in which posterior uncertainty is large because
$n_{\mathrm{tr}}\lesssim d$.

\subsection{Experiment 1: Validity and Efficiency (Low-Dimensional)}\label{sec:exp1}

\begin{figure}[ht!]
\centering
\includegraphics[width=.9\columnwidth]{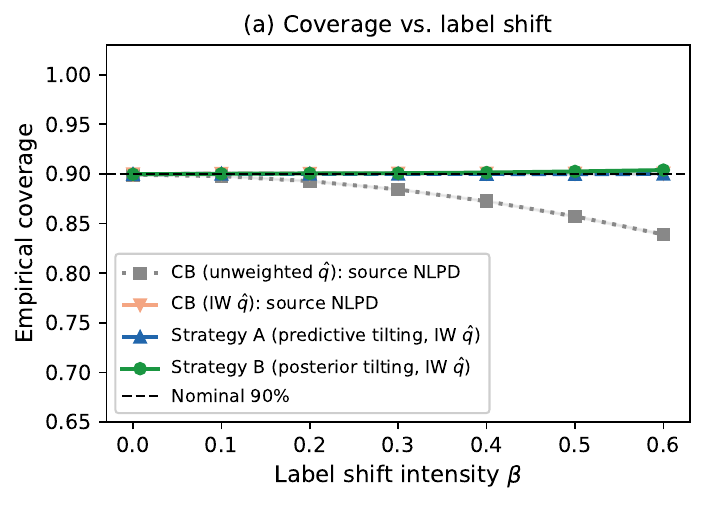}
\caption{Exp.~1: coverage vs.\ $\beta$ ($d=5$, $n_{\mathrm{tr}}=2000$).
CB (unweighted $\widehat{q}$) drops to $83.9\%$; all IW methods
maintain nominal $90\%$ coverage, confirming that validity is governed
solely by the IW quantile.}
\label{fig:cov1}
\end{figure}

Fig.~\ref{fig:cov1} confirms that empirical target coverage is governed primarily by the
IW calibration rule.
CB (unweighted $\widehat{q}$) loses coverage monotonically, reaching
$83.9\%$ at $\beta=0.6$.
All three IW methods maintain valid $90\%$ coverage throughout,
regardless of which score is used.

\begin{figure}[ht!]
\centering
\includegraphics[width=.9\columnwidth]{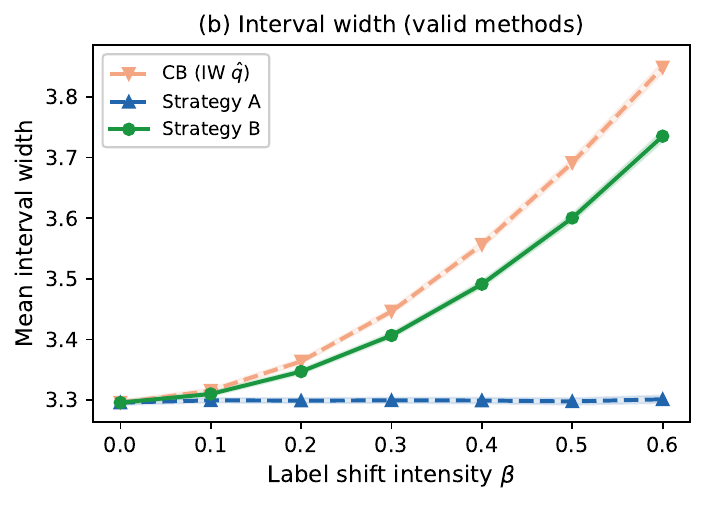}
\caption{Exp.~1: interval width vs.\ $\beta$ (valid methods).
Strategy~A is narrowest; Strategy~B is wider because the MAP
correction permanently shifts $\hth_t$ away from $\theta_{\mathrm{true}}$
in the low-dimensional, well-estimated regime.
CB (IW $\widehat{q}$) is widest.}
\label{fig:wid1}
\end{figure}

Fig.~\ref{fig:wid1} shows interval widths for the three valid methods.
Strategy~A is strictly the narrowest: with $d=5$ and $n_{\mathrm{tr}}=2000$,
the epistemic uncertainty $\hsi^2(x)$ is small and well-estimated,
so the tilted score $-\log p^A(y|x)$ is well-aligned with the true
target predictive, giving a tight IW quantile.
Strategy~B is wider because the MAP correction
$\hth_t=\hth_s-\beta n\Sigma_\theta\bar\phi$
converges to a fixed non-zero shift away from $\theta_{\mathrm{true}}$
even as $n_{\mathrm{tr}}\to\infty$, mis-centering $p^{\mathrm{B}}$
and requiring wider intervals to maintain coverage.
CB (IW $\widehat{q}$) is widest: correcting the measure without
correcting the score geometry requires a larger quantile adjustment.

\begin{remark}[When Strategy~B is suboptimal]\label{rem:subopt}
In the low-dimensional, well-estimated regime ($n_{\mathrm{tr}}\gg d$,
unbiased training), Strategy~A dominates Strategy~B on interval width
for all $\beta\neq 0$.
Strategy~B's minimum-volume property holds only when $p^{\mathrm{B}}$
is the true target predictive, which requires the parameter posterior
to be the true target posterior — a condition that holds only when
training data are drawn from $p_t$, not $p_s$.
\end{remark}

\begin{table*}[ht]
\centering
\caption{Mean absolute predictive bias
$\E_x|\hat\mu(x)-\mu_t^{\mathrm{true}}(x)|$,
Experiment~2 ($d=50$, $n_{\mathrm{tr}}=50$).
Strategy~B's bias grows slowly; Strategy~A's grows rapidly
due to large epistemic uncertainty $\hsi^2(x)\approx 6.4$.}
\label{tab:biases}
\smallskip\small
\setlength{\tabcolsep}{5pt}
\begin{tabular}{lccccccc}
\toprule
 & $\beta=0.0$ & $0.1$ & $0.2$ & $0.3$ & $0.4$ & $0.5$ & $0.6$ \\
\midrule
Strategy~A & $1.29$ & $1.36$ & $1.58$ & $1.91$ & $2.31$ & $2.77$ & $3.26$ \\
Strategy~B & $1.29$ & $1.29$ & $1.30$ & $1.31$ & $1.33$ & $1.35$ & $1.38$ \\
\bottomrule
\end{tabular}
\end{table*}

\subsection{Experiment 2: Efficiency Gain of Strategy~B (High-Dimensional)}\label{sec:exp2}

With $d=50$ and $n_{\mathrm{tr}}=50$, the posterior covariance is large:
the mean predictive variance is $\hsi^2(x)\approx 6.4$ across test inputs.
All assumptions (A1)--(A3) are satisfied — training is unbiased from
$p_s$, the model is correctly specified.

\begin{figure}[ht!]
\centering
\includegraphics[width=.9\columnwidth]{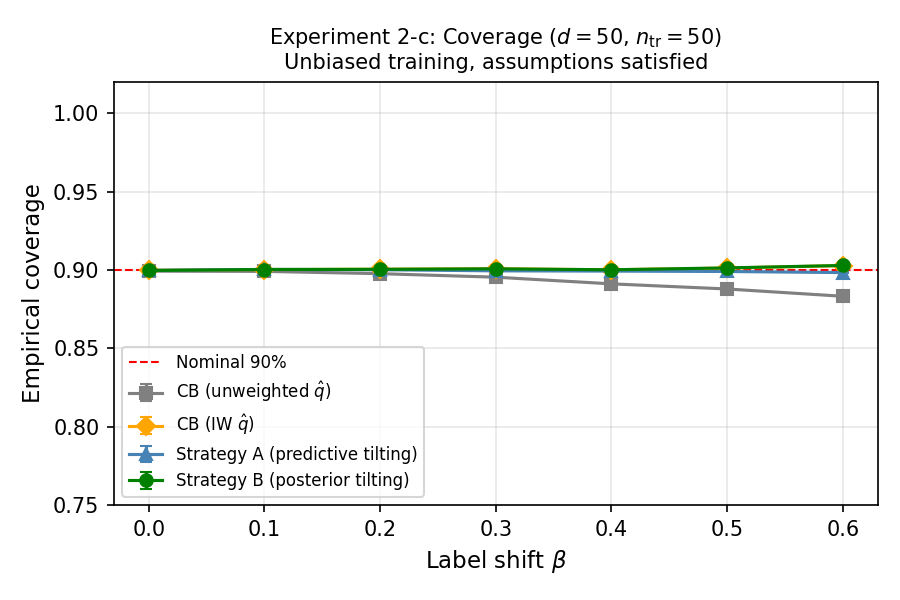}
\caption{Exp.~2: coverage vs.\ $\beta$ ($d=50$, $n_{\mathrm{tr}}=50$).
All IW methods maintain nominal $90\%$; CB (unweighted $\widehat{q}$)
degrades to $88.3\%$. Empirical target coverage is governed primarily by the IW calibration rule,
independent of dimension.}
\label{fig:cov2}
\end{figure}

Fig.~\ref{fig:cov2} confirms that all IW methods maintain valid $90\%$
coverage throughout, and CB (unweighted $\widehat{q}$) degrades to
$88.3\%$ at $\beta=0.6$.
The validity picture is identical to Experiment~1: the IW quantile
governs coverage regardless of dimension or score choice.

\begin{figure}[ht!]
\centering
\includegraphics[width=.9\columnwidth]{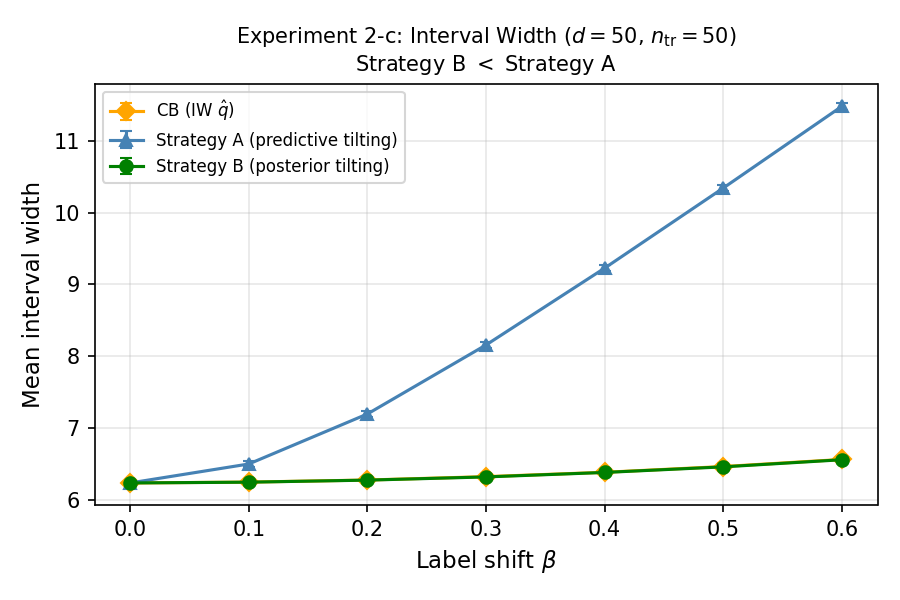}
\caption{Exp.~2: interval width vs.\ $\beta$ (valid methods, $d=50$,
$n_{\mathrm{tr}}=50$).
Strategy~B is strictly narrowest, achieving $43\%$ width reduction
over Strategy~A at $\beta=0.6$.
Strategy~A inflates dramatically because
$\beta\hsi^2(x)$ is large and heteroscedastic in this regime.}
\label{fig:wid2}
\end{figure}

Fig.~\ref{fig:wid2} shows interval widths.
The ordering is now \emph{reversed} relative to Experiment~1:
Strategy~B is strictly the narrowest, and Strategy~A is dramatically
wider.
At $\beta=0.6$, Strategy~A produces intervals of mean width $11.5$
while Strategy~B produces width $6.6$ — a \textbf{$43\%$ reduction}
at the same $90\%$ coverage.
CB (IW $\widehat{q}$) closely tracks Strategy~B throughout.

The mechanism is clear from the predictive bias (Table~\ref{tab:biases}).
Strategy~A's predicted mean shift $\beta\hsi^2(x)\approx\beta\times 6.4$
grows rapidly with $\beta$ and overshoots the true target conditional
mean $\theta_{\mathrm{true}}^\top\phi(x)+\beta\sigma^2$ for many inputs,
inflating both the calibration scores and the IW quantile threshold.
This is not a failure of the Gaussian algebra; it is the cost of tilting
the full posterior predictive when epistemic variance is large.
Strategy~B's correction $-\beta n\Sigma_\theta\bar\phi$ is a fixed
vector that shifts all predictions by a moderate, stable amount,
keeping calibration scores compact and the IW quantile low.

\begin{remark}[When Strategy~B is beneficial]\label{rem:beneficial}
In the high-dimensional, underdetermined regime ($n_{\mathrm{tr}}\lesssim d$),
the large epistemic uncertainty $\hsi^2(x)$ makes Strategy~A's tilted
score noisy and the IW quantile inflated.
Strategy~B's posterior correction avoids this: the MAP correction
shifts the predictive mean by a stable vector
$-\beta n\Sigma_\theta\bar\phi$ that is independent of $x$'s
epistemic uncertainty, keeping calibration scores compact.
The result is substantially narrower intervals at unchanged coverage,
with all theoretical assumptions (A1)--(A3) satisfied.
Strategy~B is therefore the recommended choice in high-dimensional
settings where the number of training observations is small relative
to the model dimension.
\end{remark}

\section{Limitations and Conclusion}\label{sec:disc}

We presented a conformal-Bayes framework for label shift that separates two roles:
importance-weighted calibration controls target-domain coverage, while the score
geometry determines efficiency. Strategy~A applies the shift post hoc by tilting
the posterior predictive and is preferable in the low-dimensional, well-estimated
regime. Strategy~B applies the shift at the parameter level and can be preferable
when $n_{\mathrm{tr}}\lesssim d$, where large epistemic uncertainty makes predictive
tilting overly conservative. In our controlled experiments, both strategies maintain
nominal coverage through the IW calibration rule; Strategy~A gives the narrowest
intervals in the well-estimated regime, whereas Strategy~B achieves up to $43\%$
width reduction in the high-dimensional regime at the same nominal coverage.

The framework has several limitations. It assumes pure label shift,
$p_t(x\mid y)=p_s(x\mid y)$, an exponential-tilt density-ratio model, and a known
tilt parameter $\beta$. In practice, covariate or concept shift may co-occur with
label shift, the exponential family may be misspecified, and $\beta$ must be
estimated from unlabeled target data. The present analysis is also restricted to a
Gaussian linear head, where the tilted posterior and prediction intervals are
available in closed form; extending Strategy~B to deep models requires approximate
Bayesian inference such as Laplace approximation, variational inference, Monte
Carlo dropout, or normalizing-flow approximations. Finally, the experiments are
synthetic and designed to isolate mechanisms rather than to validate performance
on real molecular data.

A practical way for estimating $\beta$ is to combine moment matching, predictive
pseudo-label sampling, and logistic density-ratio estimation, as studied in a
companion paper on molecular property prediction under label shift
\citep{LeeHS2026eiml}. The same predictive ratio also supports anytime-valid
testing for label shift via conditional e-values, developed separately in
\citet{Choi2026testing}.

\bibliography{/users/seungjin/pub/bib/sjc}
\bibliographystyle{icml2026}

\newpage
\onecolumn
\appendix

\section{Derivation of the Tilted Predictive (Strategy A)}\label{app:stratA}

We work entirely at the predictive level conditioned on $\Dtrain$,
for any fixed $\beta$.
Applying Bayes' theorem to the target distribution gives
\begin{equation}
p_t(y| x,\Dtrain)
=\frac{p_t(x| y,\Dtrain)\,p_t(y|\Dtrain)}{p_t(x|\Dtrain)}.
\label{eq:appA-step1}
\end{equation}
By (A1), $p_t(x| y,\Dtrain)=p_s(x| y,\Dtrain)$, so~\eqref{eq:appA-step1} becomes
\begin{equation}
p_t(y| x,\Dtrain)
=\frac{p_s(x| y,\Dtrain)\,p_t(y|\Dtrain)}{p_t(x|\Dtrain)}.
\label{eq:appA-step2}
\end{equation}
Applying Bayes' theorem to the source distribution,
$p_s(x| y,\Dtrain)=p_s(y| x,\Dtrain)\,p_s(x|\Dtrain)/p_s(y|\Dtrain)$,
and substituting into~\eqref{eq:appA-step2}:
\begin{equation}
p_t(y| x,\Dtrain)
=p_s(y| x,\Dtrain)\cdot
\frac{p_t(y|\Dtrain)}{p_s(y|\Dtrain)}\cdot
\frac{p_s(x|\Dtrain)}{p_t(x|\Dtrain)}.
\label{eq:appA-step4}
\end{equation}
We now identify each factor on the right.
For the label marginal ratio: since $\Dtrain$ is drawn from $p_s$,
it carries no information about the target label distribution
beyond the fixed $\beta$, so $p_t(y|\Dtrain)=p_t(y)$.
By (A3), $p_s(y|\Dtrain)\approx p_s(y)$.
Therefore
\begin{equation}
\frac{p_t(y|\Dtrain)}{p_s(y|\Dtrain)}
=\frac{p_t(y)}{p_s(y)}=w(y).
\label{eq:appA-ratio}
\end{equation}
For the covariate ratio: the factor
$p_s(x|\Dtrain)/p_t(x|\Dtrain)$ depends on $x$ but not $y$,
and equals $1/Z(x)$ where $Z(x)=\int p_s(y'| x,\Dtrain)\,w(y')\,dy'$.
To verify, we compute $p_t(x|\Dtrain)$ via the law of total
probability, using (A1) in the second equality:
\begin{align}
p_t(x|\Dtrain)
&=\int p_t(x| y',\Dtrain)\,p_t(y')\,dy' \notag\\
&=\int p_s(x| y',\Dtrain)\,p_t(y')\,dy' \notag\\
&=\int\frac{p_s(y'| x,\Dtrain)\,p_s(x|\Dtrain)}{p_s(y')}
  \,p_t(y')\,dy' \notag\\
&=p_s(x|\Dtrain)\,Z(x),
\label{eq:appA-step5}
\end{align}
so $p_s(x|\Dtrain)/p_t(x|\Dtrain)=1/Z(x)$.
Substituting~\eqref{eq:appA-ratio} and $1/Z(x)$
into~\eqref{eq:appA-step4} yields~\eqref{eq:pred-tilt}.\qed

For the Gaussian model~\eqref{eq:model} and exponential
tilting~\eqref{eq:tilt}, the source predictive is
$p_s(y|x,\Dtrain)=\mathcal{N}(y;\hmu_s(x),\hsi^2(x))$.
Computing $Z(x)=\int\mathcal{N}(y';\hmu_s(x),\hsi^2(x))\,w(y')\,dy'$
by completing the square gives
\[
Z(x)=\exp\!\bigl(\beta\hmu_s(x)+\tfrac{1}{2}\beta^2\hsi^2(x)\bigr)/Z_w,
\]
so $p^{\mathrm{A}}(y|x,\Dtrain)=\mathcal{N}(y;\hmu_s(x)+\beta\hsi^2(x),\hsi^2(x))$
and the score~\eqref{eq:sA} is its NLPD.\qed

\section{Derivation of the Corrected Posterior (Strategy B)}\label{app:stratB}

Since the backbone $\phi$ is frozen, $p(x)$ does not depend on
$\theta$ and the source joint factorizes as
$p_s(x,y|\theta)=p_s(y|x,\theta)\,p(x)$.
Under (A1)--(A2), the target joint is obtained by reweighting by
$w(y)=p_t(y)/p_s(y)$:
\[
p_t(x,y|\theta)
\;\propto\; p_s(y|x,\theta)\,p(x)\,w(y).
\]
Integrating out $y$ gives the target covariate marginal
$p_t(x|\theta)=p(x)\,Z_\theta(x)$,
where $Z_\theta(x)=\int p_s(y'|x,\theta)\,w(y')\,dy'$.
Dividing the target joint by this marginal, $p(x)$ cancels, giving
the $\theta$-dependent likelihood ratio:
\begin{equation}
\frac{p_t(y|x,\theta)}{p_s(y|x,\theta)}
=\frac{w(y)}{Z_\theta(x)},
\qquad
Z_\theta(x)=\int p_s(y'|x,\theta)\,w(y')\,dy'.
\label{eq:lr}
\end{equation}

Applying~\eqref{eq:lr} to each training observation, the target
posterior is
\[
\pi_t(\theta|\Dtrain)
\;\propto\;
\pi_s(\theta|\Dtrain)\cdot\prod_{i=1}^n
\frac{w(y_i)}{Z_\theta(x_i)}.
\]
Since $\prod_i w(y_i)$ does not depend on $\theta$, it is absorbed
into the normalization constant, leaving:
\begin{equation}
\pi_t(\theta|\Dtrain)
\;\propto\;
\pi_s(\theta|\Dtrain)\cdot\prod_{i=1}^n Z_\theta(x_i)^{-1}.
\label{eq:post-t-app}
\end{equation}

For the Gaussian model~\eqref{eq:model} and exponential
tilting~\eqref{eq:tilt} with $w(y)=\exp(\beta y)/Z_w$,
completing the square in the exponent of
$Z_\theta(x)=Z_w^{-1}\int\mathcal{N}(y';\theta^\top\phi(x),\sigma^2)
\exp(\beta y')\,dy'$ gives:
\begin{equation}
Z_\theta(x)
=\frac{1}{Z_w}\exp\!\bigl(\beta\theta^\top\phi(x)
  +\tfrac{1}{2}\beta^2\sigma^2\bigr),
\label{eq:Ztheta}
\end{equation}
so $\log Z_\theta(x_i)=\beta\,\theta^\top\phi(x_i)+\mathrm{const}$
with a $\theta$-free constant.
Substituting~\eqref{eq:Ztheta} into~\eqref{eq:lr} also gives the
Gaussian target conditional
$p_t(y|x,\theta)=\mathcal{N}(y;\theta^\top\phi(x)+\beta\sigma^2,\sigma^2)$.

Substituting $\log Z_\theta(x_i)=\beta\,\theta^\top\phi(x_i)+\mathrm{const}$
into~\eqref{eq:post-t-app}, the log-posterior becomes
\[
\log\pi_t(\theta|\Dtrain)
=\log\pi_s(\theta|\Dtrain)
 -\beta\sum_{i=1}^n\theta^\top\phi(x_i)+\mathrm{const}.
\]
The correction is linear in $\theta$, so setting
$\nabla_\theta\log\pi_t(\theta|\Dtrain)=0$ gives
\[
-\Sigma_\theta^{-1}(\theta-\hth_s)
-\beta\sum_{i=1}^n\phi(x_i)=0,
\]
and hence
$\hth_t=\hth_s-\beta n\Sigma_\theta\bar\phi$
where $\bar\phi=n^{-1}\sum_{i=1}^n\phi(x_i)$.
The Hessian $-\nabla^2_\theta\log\pi_t(\theta|\Dtrain)=\Sigma_\theta^{-1}$
is unchanged (a linear correction contributes zero to the second
derivative), so $\widehat\Sigma_t=\Sigma_\theta$.\qed

\end{document}